\documentclass{article}
\usepackage{spconf}
\usepackage{amsmath}
\usepackage{graphicx}
\usepackage{amssymb}
\usepackage{fancyhdr}
\usepackage[OT1]{fontenc}
\fancypagestyle{alim}{
    \fancyhf{}
    
    \fancyfoot[c]{{\small © 2020 IEEE.  Personal use of this material is permitted.  Permission from IEEE must be obtained for all other uses, in any current or future media, including reprinting/republishing this material for advertising or promotional purposes, creating new collective works, for resale or redistribution to servers or lists, or reuse of any copyrighted component of this work in other works.}}
}

\title{Retrieving and Highlighting Action with Spatiotemporal Reference}
\name{
    \begin{tabular}{c}
    Seito Kasai$^{\star}$ \qquad
    Yuchi Ishikawa$^{\star}$ \qquad
    Masaki Hayashi$^{\dagger}$ \\
    Yoshimitsu Aoki$^{\dagger}$ \qquad
    Kensho Hara$^{\star}$ \qquad
    Hirokatsu Kataoka$^{\star}$
    \end{tabular}
}
  
\address{$^{\star}$ AIST \\
$^{\dagger}$ Keio University}
\begin{document}
\thispagestyle{alim}
\maketitle
\begin{abstract}
In this paper, we present a framework that jointly retrieves and spatiotemporally highlights actions in videos by enhancing current deep cross-modal retrieval methods.
Our work takes on the novel task of action highlighting, which visualizes where and when actions occur in an untrimmed video setting.
Action highlighting is a fine-grained task, compared to conventional action recognition tasks which focus on classification or window-based localization.
Leveraging weak supervision from annotated captions, our framework acquires spatiotemporal relevance maps and generates local embeddings which relate to the nouns and verbs in captions.
Through experiments, we show that our model generates various maps conditioned on different actions, in which conventional visual reasoning methods only go as far as to show a single deterministic saliency map.
Also, our model improves retrieval recall over our baseline without alignment by 2-3\% on the MSR-VTT dataset.
\end{abstract}


\section{Introduction} \label{sec:intro}
Given the exponential growth of media in the digital age, precise and fine-grained retrieval becomes a crucial issue.
Content-based search, specifically embedding learning, is done for media retrieval within large amounts of data \cite{vse, vsec, vsepp, adv_rep, dualencoding, univse, jpose, beansinburgers, moee, multimodalcues, gxn}.
Unlike simple classification from predefined labels, these methods learn instance-specific embeddings, allowing finer search granularity.
Using embedding-based methods, retrieval is accomplished efficiently even in the cross-modal setting by carrying out a nearest neighbor search in the embedding space.

Embedding learning and retrieval of videos from text is especially challenging, due to the fact that videos contain substantial low-level information compared to semantically abstract captions.
That is, video captions include high-level, global information often describing either only a salient part of a video or what the whole video is about.
Conventional approaches to this problem in embedding learning between video and text leverage various rich features from videos which are not limited to visual phenomena, such as optical flow, sound, and detected objects \cite{moee, usewhatyou}.
This aids the extraction of high-level information crucial for making caption-like embeddings.
However, model interpretability degrades as more features are generated, and the reason behind the retrieval result is ambiguous.

To alleviate the gap in abstractness and adhere interpretability when retrieving videos from captions, we propose the novel task of action highlighting.
Action highlighting is a challenging task for generating spatiotemporal voxels which score the relevance of the local region to a certain action class or word.
For this problem, our method generates these voxels in a weakly-supervised manner, given only pairs of videos and captions.
Aligning local regions extracted by the features from a spatiotemporal 3D-CNN to nouns and verbs in captions, our model generates robust embeddings for both videos and associated captions, which can then be used for retrieval.

\begin{figure}
\begin{center}
\includegraphics[width=1.0\linewidth]{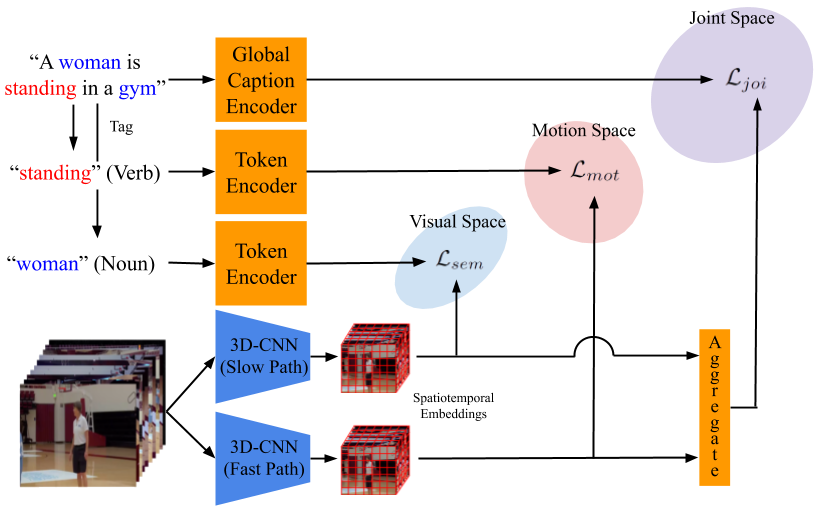}
\end{center}
   \caption{
   Overview of our proposed framework.
   We use different encoders for captions and parsed tokens.
   In terms of videos, we use two different branches for extracting slow and fast features, retaining spatiotemporal information.
   }
\label{fig:mse_space}
\end{figure}

We hypothesize that the information in captions that is crucial to retrieval concerns ``what'' is done ``when'' and ``where'' in the video.
Therefore, as shown in Figure~\ref{fig:mse_space}, we learn three different embeddings for the video-caption retrieval problem, namely the motion, visual, and joint spaces.
These three spaces model the embeddings for verbs, nouns and whole captions respectively, to extract specific features from videos which relate to the above mentioned crucial information of the caption.
This aids the model in obtaining important motion and visual features from the video, alleviating the redundancy existent in video features compared to caption features.

The contributions of our work are two-fold;
\setlength{\itemsep}{0cm}
\begin{itemize}
    \item We address the novel setting of action highlighting, which requires the generation of spatiotemporal voxels conditioned on action words, indicating where and when actions occur.
    This enables reasoning and model interpretability in video-text retrieval.
    \item We show through experiments that our spatiotemporal alignment loss generates local embeddings associated to verbs or nouns in captions, and improves video-text retrieval performance by 2-3\% compared to our baseline model.
\end{itemize}

\section{Method}
\label{sec:method}

\subsection{Feature Extraction} \label{sec:ft_ext}
To extract motion and visual features from videos, we use the SlowFast Network \cite{slowfast}.
This network uses two different branches, with the slow branch extracting more visual information using higher spatial resolution, and the fast branch focusing on high-framerate motion information.
The feature from the slow branch, $\mathbf{v}_{slow} \in \mathbb{R}^{C_s \times T_s \times H \times W}$, will be projected to the visual space and that of the fast branch, $\mathbf{v}_{fast} \in \mathbb{R}^{C_f \times T_f \times H \times W}$, will be projected to the motion space.
Here $C_s$ and $C_f$ are the channel dimensions, $T_s$ and $T_f$ are the temporal dimensions for the features extracted by the slow and fast paths respectively.

For textual features, we first tagged the captions and extracted the tokens that have a part-of-speech tag with verbs and nouns.
These verb and noun tokens are converted to distributed representations with an learnable embedding matrix to generate vectors $\mathbf{t}_{vb}, \mathbf{t}_{nn} \in \mathbb{R}^{E}$.
Then, we encode these nouns and verbs with a token encoder, which is a GRU cell to project these features to each of the motion and visual spaces:
\begin{align} \label{eq:gru}
    \mathbf{c}_{mot}^+ &= \sigma(\mathbf{W}_1\mathbf{t}_{vb} + \mathbf{b}_1) \tanh(\mathbf{W}_2\mathbf{t}_{vb} + \mathbf{b}_2) \\
    \mathbf{c}_{vis}^+ &= \sigma(\mathbf{W}_3\mathbf{t}_{nn} + \mathbf{b}_3) \tanh(\mathbf{W}_4\mathbf{t}_{nn} + \mathbf{b}_4), 
\end{align} \noindent
where $\sigma$ denotes the sigmoid function, and $\mathbf{W} \in \mathbb{R}^{E \times C}, \mathbf{b} \in \mathbb{R}^{C}$ are weights and biases for affine transformation.
We also use the sets of all verbs and nouns in the whole training dataset to sample the negative verbs and nouns that are not seen in the same caption sets, and use the same model to generate $\mathbf{c}^-_{mot}, \mathbf{c}^-_{vis} \in \mathbb{R}^C$.

We also extract global textual features using a caption encoder, which is a simple GRU.
We denote this feature as $\mathbf{c}^+_{joi}$.
Whole negative caption features that are not in the same caption set are randomly sampled, and their features are denoted $\mathbf{c}^-_{joi}$.
In our implementation, all textual embeddings generated from the caption and token encoders are normalized via $L^2$.

\begin{figure}[t]
\begin{center}
\includegraphics[width=0.9\linewidth]{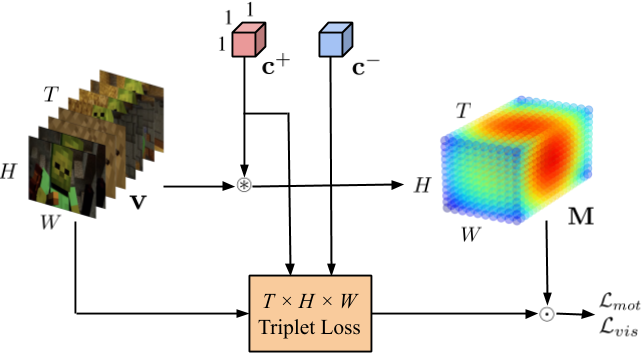}
\end{center}
   \caption{Motion-Visual alignment loss details.
   $\odot$ denotes the element-wise product, and $\circledast$ denotes convolution of stride 1.
   $T = T_f$ for $\mathcal{L}_{mot}$ and $T = T_s$ for $\mathcal{L}_{vis}$.
   }
\label{fig:align}
\end{figure}

\subsection{Motion-Visual Alignment} \label{sec:method2}
Here, we describe our novel method on how to generate the spatiotemporal relevance map for action highlighting and align nouns and verbs to local regions in the spatiotemporal embeddings.

First, we pass SlowFast features $\mathbf{v}_{fast}, \mathbf{v}_{slow}$ through a single trainable 3D convolution of kernel size 1, and project them to $\mathbf{v}_{mot}, \mathbf{v}_{vis} \in \mathbb{R}^{C \times T \times H \times W}$ respectively.
Note that the spatiotemporal dimension sizes $(T, H, W)$ are arbitrary.
Figure~\ref{fig:align} shows our feature space alignment for both motion and visual spaces.
Given the $(i, j, k)$-th position of our video feature vector $\mathbf{v}^{ijk} \in \mathbb{R}^C$, we use the positive token encodings, $\mathbf{c}^+_{mot}$ and $\mathbf{c}^+_{vis}$, to calculate relevance maps $\mathbf{M} \in \mathbb{R}^{T \times H \times W}$ in an attention-like procedure.
The relevance map's $(i, j, k)$-th element is calculated by
\begin{equation} \label{eq:relmap}
    m^{ijk}_l = \frac{e^{\,s(\mathbf{v}^{ijk}_l, \mathbf{c}^+_l) / \beta}}{\sum_{ijk}{e^{\,s(\mathbf{v}^{ijk}_l, \mathbf{c}^+_l) / \beta}}},
    \hspace{10pt} l \in \{mot, vis\},
\end{equation} \noindent
where $s(., .)$ denotes the similarity function, implemented as cosine similarity.
Note that the softmax temperature $\beta$ determines how ``sharp'' the relevance map would be, and can be modified during inference for visualization.
Then, we apply a weighted $T \times H \times W$ triplet loss function for each of the spatiotemporal positions of the video feature vector as per the following equations,
\begin{equation} \label{eq:loss_align}
    \mathcal{L}_{l} = \underset{ijk}{\sum}\ m^{ijk}_{l} \big\lfloor \alpha - s(\mathbf{v}^{ijk}_{l}, \mathbf{c}^+_{l}) + s(\mathbf{v}^{ijk}_{l}, \mathbf{c}^-_{l})\big\rfloor_+ , \nonumber \\
    l \in \{mot, vis\} .
\end{equation} \noindent
Here, $\lfloor \cdot \rfloor_+$ is equivalent to $\max(0,\cdot)$, and $\alpha$ is the margin hyperparameter.
In this triplet loss function, we only use the negatives for the caption side ($\mathbf{c}^-$), rendering it unidirectional.
This weighting with the relevance map $\mathbf{M}$ allows the alignment loss to only be computed locally where textual features describing motion or visual information are close to visual features.

We calculate this loss between the embeddings of video features from the fast branch, $\mathbf{v}_{mot}$ with the textual features of the nouns $\mathbf{c}_{mot}$, and the embeddings of video features from the slow branch $\mathbf{v}_{vis}$ with the textual features of the verbs $\mathbf{c}_{vis}$, resulting in two relevance maps $\mathbf{M}_{mot}$ and $\mathbf{M}_{vis}$ as well as two loss terms $\mathcal{L}_{mot}$ and $\mathcal{L}_{vis}$.

\subsection{Joint Space Learning} \label{sec:method3}
Using the video features $\mathbf{v}_{mot}$ and $\mathbf{v}_{vis}$, we construct a global feature for the video.
We simply concatenate and project the two features using a single $(1 \times 1 \times 1)$ 3d convolution with sigmoid activation into a joint video feature,
\begin{equation} \label{eq:joint}
    \mathbf{v}_{joi} = L^2\,\text{Norm}\,(\sigma(\mathbf{W}[\mathbf{v}_{mot},\mathbf{v}_{vis}] + \mathbf{b})),
\end{equation} \noindent
where $\mathbf{W} \in \mathbb{R}^{2C \times C}$ and $\mathbf{b} \in \mathbb{R}^C$.

The joint embedding loss, $\mathcal{L}_{joi}$ is calculated as
\begin{align} \label{eq:loss_joint}
    \mathcal{L}_{joi} = \mathbb{E}_{\mathcal{B}} \Big[\, \big\lfloor \alpha - s(\mathbf{v}^+_{joi},\ \mathbf{c}^+_{joi}) + s(\mathbf{v}^+_{joi},\ \mathbf{c}^-_{joi}) \big\rfloor_+ \nonumber \\
    + \big\lfloor \alpha - s(\mathbf{v}^+_{joi},\ \mathbf{c}^+_{joi}) + s(\mathbf{v}^-_{joi},\ \mathbf{c}^+_{joi}) \big\rfloor_+ \,\Big] ,
\end{align} \noindent
where $\mathcal{B}$ denotes a batch and the $(+, -)$ on the shoulder of vectors denote if the sample is a positive or a negative.
This objective is known as the triplet loss \cite{facenet} and, specifically, we use all of the negatives in a single batch for good gradient signals during training following \cite{multimodalcues}.

\subsection{Training Objective, Retrieval and Reranking} \label{sec:method5}
Our overall loss function for training embeddings, $\mathcal{L}_{all}$ is
\begin{equation} \label{eq:loss_all}
    \mathcal{L}_{all} = \mathcal{L}_{joi} + \lambda_{m} \mathcal{L}_{mot} + \lambda_{s} \mathcal{L}_{vis}.
\end{equation} \noindent

For cross-modal retrieval, we evaluate $s(\mathbf{v}^m_{joi}, \mathbf{c}^n_{joi})$ for each $m$-th video and $n$-th caption to get similarity rankings for all combinations of embeddings.

Additionally, we use the motion and visual spaces to conduct reranking and further refine our retrieval.
Tagging the tokens of the captions, we get the feature vectors of the verbs and nouns $\mathbf{c}_{mot}$ and $\mathbf{c}_{vis}$.
Then, we pool the features $\mathbf{v}^{ijk}_{mot}$ and $\mathbf{v}^{ijk}_{vis}$ to get global motion and visual video representations:
\begin{equation} \label{eq:pool}
    \hat{\mathbf{v}}_{l} = \frac{1}{T_fHW} \underset{ijk}{\sum}\, \mathbf{v}^{ijk}_{l} , \hspace{10pt} l \in \{mot, vis\} .
\end{equation} \noindent
Evaluating $s(\hat{\mathbf{v}}_{mot}^m, \mathbf{c}^n_{mot})$ and $s(\hat{\mathbf{v}}_{vis}^m, \mathbf{c}^n_{vis})$, we conduct reranking by summing these similarities in the embedding spaces before building the rankings.

\section{Experiments and discussions}

We use the MSR-VTT \cite{msrvtt} dataset for our retrieval experiments, which consists of 10k short but untrimmed videos.
Additionally, we use part of the Kinetics-700 dataset \cite{kinetics} for action highlighting, which consists of 700 action classes with 650K trimmed videos in total.

\begin{figure}[t]
\begin{center}
\includegraphics[width=1.0\linewidth]{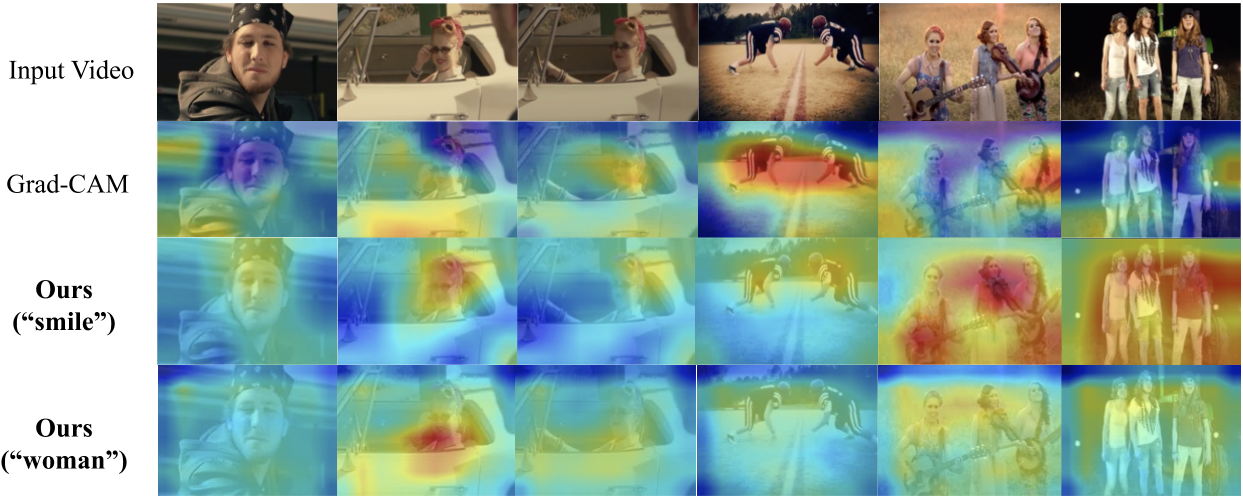}
\end{center}
   \caption{
   Action highlighting on MSR-VTT, compared with Grad-CAM \cite{gradcam} visualization results.
   Note that this is not a direct comparison as Grad-CAM shows reasoning for its output (in this example, generated captions), while ours shows highlighting with respect to the token of interest (in braces).
   }
\label{fig:highlight}
\end{figure}

\begin{figure}[t]
\begin{center}
\includegraphics[width=1.0\linewidth]{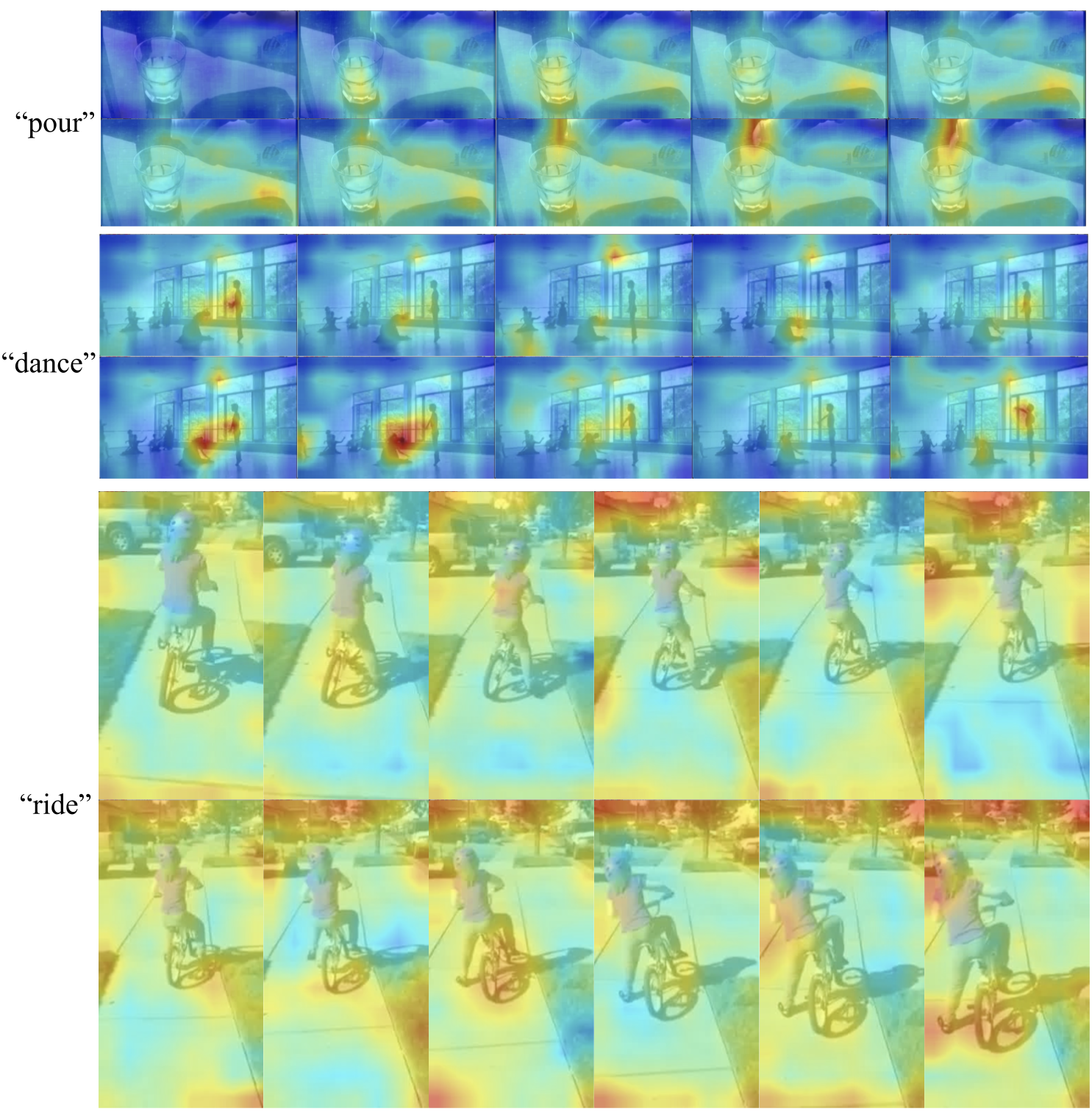}
\end{center}
   \caption{
   Action highlighting on the Kinetics dataset.
   The top row highlights ``pour'' in the class ``pouring milk'', the middle highlights ``dance'' in the class ``dancing ballet'', and the bottom highlights ``ride'' in the class ``riding a bike''.
   }
\label{fig:highlight_k}
\end{figure}

\subsection{Comparison with Grad-CAM} \label{sec:highlighting}
In Figure~\ref{fig:highlight}, we compare our results of action highlighting in the MSR-VTT dataset to Grad-CAM \cite{gradcam} saliency visualization.
For the gradient signals used in Grad-CAM visualization, we train a simple video captioning model using a 3D-ResNet \cite{3dresnet} encoder and an LSTM decoder.
However, Grad-CAM shows which pixels contributed to the output of the downstream task, video captioning in this case, and is therefore insufficient for action highlighting.
For our results, we show highlighting results done on the same video conditioned on two words ``smile'' and ``woman'', each for the motion and visual spaces.

Note that, during inference, the visualized regions using Grad-CAM are deterministic.
On the contrary, ours notably generates different maps for different tokens even for a single video, and produces verb-conditioned action highlighting.
Moreover, general search queries use the open vocabulary, which our model covers.

\subsection{Highlighting Actions in the Kinetics Dataset} \label{sec:highlighting_k}
See Figure~\ref{fig:highlight_k} for results of action highlighting using the Kinetics-700 dataset.
Of the 700 action class labels introduced in the dataset, we use a small subset and extract nouns and verbs in the action class names, using them to show relevance maps for both the motion and visual spaces.
Because of the fully convolutional nature of our model, we were able to extract relevance maps with varying resolution and aspect ratio.
This is crucial in action highlighting for natural videos, as the spatiotemporal resolution is not distorted with respect to the resizing of the frames.

From the first row in Figure~\ref{fig:highlight_k}, the responsive local features to ``pour'' include the poured milk as well as the target cup, thus showing understanding of which local region represents the action well.
The second row shows two blobs of local embeddings similar to the representation of the token ``dance''.
We point out that strong responses do not span over the whole silhouette of the people dancing, but only moving body parts such as the hand or head.

From the above results, our model shows to ground high-level features into local regions that represent the action, not superficial features that depend on the object.
On the contrary, the third row shows a failure case for the token ``ride''.
In this scene, our model is expected to highlight the child ``riding'' the bicycle, but instead embeds the surrounding regions close to ``ride''.
Since videos with people ``riding'' often show similar surroundings, for example a scene showing a street with trees, we believe that our model attempting to encode high-level information misunderstands the semantics of these verbs.

\begin{table}
\begin{center}
\begin{tabular}{ccc||cc|cc}
\hline
\multicolumn{3}{c}{Components} & \multicolumn{2}{c}{Video2Caption} & \multicolumn{2}{c}{Caption2Video} \\
\hline
SF & $\mathcal{L}_{mot}$ & $\mathcal{L}_{vis}$ & R@1 & Med r & R@1 & Med r \\
\hline
           &            &            & 2.6 & 74.5 & 2.4 & 57.0 \\
\checkmark &            &            & 3.6 & 45.5 & 3.1 & 42.0 \\
\checkmark & \checkmark &            & 3.6 & 52.5 & 3.3 & 48.0 \\
\checkmark &            & \checkmark & 4.9 & 55.0 & 3.1 & 46.0 \\
\checkmark & \checkmark & \checkmark & \textbf{5.2} & \textbf{42.0} & \textbf{5.3} & \textbf{40.0} \\
\hline
\end{tabular}
\end{center}
\caption{
Ablation study with model components.
SF denotes using SlowFast Networks \cite{slowfast} for feature extraction, and the top row uses 3D-ResNet \cite{3dresnet} features as an alternative.
R@1 denotes the percentage of positives within the top candidates.
Med r denotes the median ranking of highest ranked positives. 
}
\label{tab:ablation}
\end{table}

\subsection{Ablation Studies} \label{sec:ablation}
In Table \ref{tab:ablation}, we evaluate the effect of our model components.

The first and second rows show retrieval recall and median ranking when training with 3D-ResNet features and SlowFast features respectively.
Note that simply using the slow and fast features for training improves the retrieval performance substantively, suggesting that SlowFast features are richer and provide better signals for cross-modal matching.

From the second and third rows, we can see that the motion space alignment itself shows marginal improvement of the score.
The motion space alignment is difficult compared to visual space alignment, due to the fact that local regions in the video do not correlate strongly with verbs in the caption.
On the contrary, the second and fourth rows show a noticeable difference in recall for the video-to-caption task.
By aligning the visual space with nouns, or tokens that refer to objects or people in the video, the model learns to generate object-aware local embeddings.
Finally, through the fifth row, we see that a combination of verb and noun alignment improves the Recall@1 by 2-3\% for both retrieval directions.
Therefore, motion and visual alignments, especially when done altogether, show to provide effective signals when generating cross-modal embeddings.

\section{Conclusion}
In this work, we proposed the novel task of action highlighting, which requires the generation of spatiotemporal maps where and when an action occurs.
Action highlighting is a more fine-grained task for retrieving actions from videos compared to conventional action recognition tasks.

Using pairs of videos and captions, our proposed model aligns spatiotemporal local embeddings to nouns and verbs in captions, which references which part of a video represents the desired action.
Visualizing where these maps highlight a video, our method incorporates interpretability in the video-text retrieval task.
Empirical results show that our generated local embeddings go farther than simple object representations, and encode high-level information which enhances cross-modal retrieval performance as well.

\bibliographystyle{IEEEbib}
\bibliography{strings,refs}

\end{document}